\documentclass{article}
\usepackage{spconf,amsmath,epsfig}
\usepackage{amsmath, amssymb}
\usepackage[numbers]{natbib}
\usepackage{booktabs}
\usepackage{lineno}
\usepackage[linesnumbered,boxed,ruled,commentsnumbered]{algorithm2e}
\usepackage{xcolor}
\usepackage{hhline}
\usepackage{booktabs}
\usepackage{tabularx}
\usepackage{graphicx}
\usepackage{pifont}
\usepackage{mathrsfs}

\newcolumntype{Y}{>{\centering\arraybackslash}X}

\title{SAR DESPECKLING VIA REGIONAL DENOISING DIFFUSION PROBABILISTIC MODEL}
\name{Xuran Hu\textsuperscript{1}, Ziqiang Xu\textsuperscript{1, 2}, Zhihan Chen\textsuperscript{1}, Zhenpeng Feng\textsuperscript{1}, Mingzhe Zhu\textsuperscript{1, 2}, Ljubi\v{s}a Stankovi\'c\textsuperscript{3}}
\address{\textsuperscript{1} School of Electronic Engineering, Xidian University, China \\ \textsuperscript{2} Kunshan Innovation Institute of Xidian University, China \\ \textsuperscript{3} Faculty of Electrical Engineering, University of Montenegro, Montenegro}

\begin{document}
\topmargin=0mm
%
\maketitle
\begin{abstract}
	
Speckle noise poses a significant challenge in maintaining the quality of synthetic aperture radar (SAR) images, so SAR despeckling techniques have drawn increasing attention. Despite the tremendous advancements of deep learning in fixed-scale SAR image despeckling, these methods still struggle to deal with large-scale SAR images. To address this problem, this paper introduces a novel despeckling approach termed Region Denoising Diffusion Probabilistic Model (R-DDPM) based on generative models. R-DDPM enables versatile despeckling of SAR images across various scales, accomplished within a single training session. Moreover, The artifacts in the fused SAR images can be avoided effectively with the utilization of region-guided inverse sampling. Experiments of our proposed R-DDPM on Sentinel-1 data demonstrates superior performance to existing methods.

\end{abstract}
\section{Introduction}
\label{sec:intro}

Synthetic Aperture Radar (SAR) has evolved into an indispensable technology utilized across diverse domains, including remote sensing, electronic reconnaissance, and disaster response. Nonetheless, SAR imagery frequently encounters a persistent challenge known as speckle noise, stemming from the superposition of coherent radar echoes. Consequently, there has been a notable surge of interest in developing despeckling techniques within SAR image processing.

Deep-learning-based SAR despeckling methods recently obtained tremendous success in SAR despeckling. SAR-CNN \cite{chierchia2017sar, tucker2022polarimetric} employ residual learning, significantly enhancing convergence speed. InSAR-MONet \cite{vitale2022insar} proposed a new loss function to analyze the image's statistical properties. SAR2SAR \cite{dalsasso2021sar2sar} and MERLIN \cite{dalsasso2021if} integrate considerations on the spatial correlation intrinsic to SAR data.

Generative models, as a specific type of deep networks, have found widespread application in the field of SAR image restoration. With the introduction of conditional generative models like conditional Generative Adversarial Networks (c-GAN) \cite{mirza2014conditional} and Diffusion Denoising Probabilistic Model (DDPM) \cite{ho2020denoising, nichol2021improved, dhariwal2021diffusion}, novel approaches for SAR despeckling have emerged \cite{perera2023sar, xiao2023unsupervised}. DDPM employs an iterative diffusion process to progressively reduce image noise, resulting in clear, high-fidelity images. Differing from conventional generation methods, diffusion refines the image despeckling process gradually and continuously by diffusing and sampling the noise. This gradual approach allows the model to retain the image's details and structure while minimizing the impact of noise, providing a new avenue for SAR despeckling. However, these methods remain preliminary in SAR image restoration and lack generalizability for large-scale SAR images despeckling. Specifically, the structure of DDPM leads to an exponential surge in computing resource requirements with increasing image sizes. Considering the typically extensive sampling ranges in SAR images (e.g., Sentinel-1), the current diffusion-based method \cite{perera2023sar, xiao2023unsupervised, guha2023sddpm} needs to catch up to large-scale SAR image despeckling tasks. However, block-wise processing leads to the occurrence of edge artifacts.

To address above issues, this paper introduces a novel SAR despeckling diffusion model termed the Regional Diffusion Denoising Probability Model (R-DDPM), designed to address SAR image speckle noise effectively at any scale. Specifically, the diffusion process involves training the Diffusion model through image blockization, while the reverse process ensures smooth edge sampling of blocks through a guided denoising procedure \cite{ozdenizci2023restoring}. Simultaneously, we incorporate DDIM \cite{song2020denoising} to streamline the sampling inference steps, enabling faster SAR image despeckling. The primary contributions of this paper are as follows: (1) We propose regional diffusion, which achieves SAR despeckling at arbitrary scales with a single training process; (2) We introduce a novel lightweight model to minimize the deployment requirements of Diffusion in SAR despeckling; (3) Our approach demonstrates superior performance on Sentinel-1 data to existing methods.

\section{Methodology}
\label{section:B}

\subsection{Dinoising Diffusion Probabilistic Model}

Denoising Diffusion Probabilistic Model \cite{ho2020denoising, nichol2021improved} belong to the category of generative models rooted in Markov processes. The primary aim of DDPM is to forecast the target distribution $x_0 \sim q(x)$ from a Gaussian distribution $x_{T \rightarrow \infty} \sim \mathcal{N}(0,1)$ via deep learning. The implementation of DDPM involves two distinct stages: the diffusion process and the reverse diffusion process. During the diffusion process, Gaussian noise is systematically introduced into the initial distribution $x_0$, gradually transitioning it into an intermediate state $x_t$. This iterative process continues until the original distribution ultimately converges to a Gaussian distribution $x_T$:

\begin{figure}[!t]
	\centering
	\includegraphics[width=\linewidth]{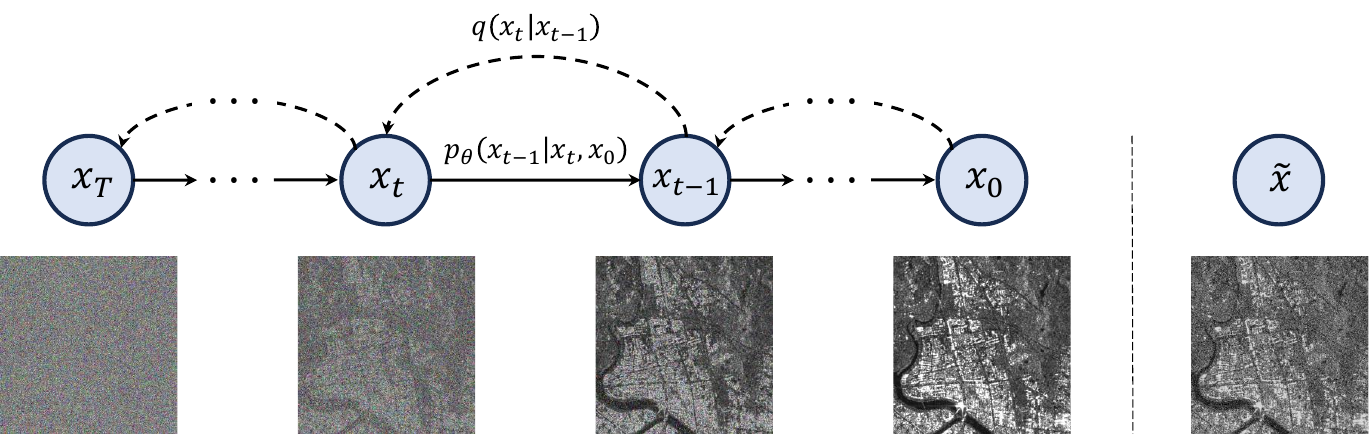}
	\caption{An overview of the diffusion and reverse processes of conditional diffusion model.}
	\label{fig_1}
\end{figure}

\begin{figure}[!t]
	\centering
	\includegraphics[width=\linewidth]{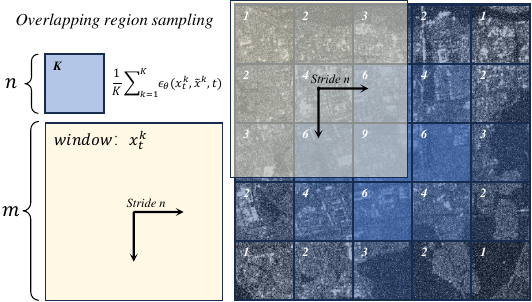}
	\caption{Illustration of regional restoration: diffusion sampling involves moving the 'window' $x_t^k$ across diverse regions, culminating in the final despeckled outcome by averaging overlapping regions.}
	\label{fig_2}
\end{figure}

\begin{equation}
	\small
	\begin{gathered}
		q\left(x_t \mid x_{t-1}\right)=\mathcal{N}\left(x_t ; \sqrt{1-\beta_t} x_{t-1}, \beta_t I\right) \\
		q\left(x_{1: T} \mid x_0\right)=\prod_{t=1}^T q\left(x_t \mid x_{t-1}\right) \\
		q\left(x_t \mid x_0\right)=\mathcal{N}\left(x_t ; \sqrt{\prod_{i=1}^T\left(1-\beta_t\right)} x_0,\left(1-\prod_{i=1}^T\left(1-\beta_t\right)\right) I\right)
	\end{gathered}
\end{equation}

Among these parameters, $\beta_t$ represents a hyperparameter, necessitating the condition $(i>j) \Rightarrow\left(\beta_i>\beta_j\right)$, and $I \sim \mathcal{N}(0,1)$. Let $\alpha_t=1-\beta_t, \overline{\alpha_t}=\prod_{i=1}^T \alpha_t$. Equation (3) can thus be reformulated as:

\begin{equation}
	\small
	q\left(x_t \mid x_0\right)=\mathcal{N}\left(x_t ; \sqrt{\overline{\alpha_t}} x_0,\left(1-\overline{\alpha_t}\right) I\right)
\end{equation}

The reverse diffusion process progressively anticipates the target distribution starting from the Gaussian distribution. When $\beta_t$ is sufficiently small, each interim step within the reverse diffusion process approximates a Gaussian distribution. Consequently, the primary objective revolves around training the network model $p_\theta$ to accurately estimate the conditional probability during this inverse process:

\begin{equation}
	\small
	\begin{gathered}
		p_\theta\left(x_{t-1} \mid x_t\right)=\mathcal{N}\left(x_{t-1} ; \mu_\theta\left(x_t, t\right), \sigma_t I\right) \\
		p_\theta\left(x_{0: T}\right)=p\left(x_T\right) \prod_{t=1}^T p\left(x_{t-1} \mid x_t\right)
	\end{gathered}
\end{equation} where the variance $\sigma_t^2$ is set to our chosen value $\beta_t$. Than we compute the posterior conditional probability $q\left(x_{t-1} \mid x_t, x_0\right)$ during the diffusion process and derive its mean and variance:

\begin{equation}
	\small
	\begin{gathered}
		q\left(x_{t-1} \mid x_t, x_0\right)=\mathcal{N}\left(x_{t-1} ; \tilde{\mu}\left(x_t, x_0\right), \widetilde{\beta}_t I\right) \\
		\propto \exp \left\{-\frac{1}{2}\left[\left(\frac{\alpha_t}{\beta_t}+\frac{1}{1-\bar{\alpha}_{t-1}}\right) x_{t-1}^2\right)\right. \\
		\left.\left.\left.\quad-\left(\frac{2 \sqrt{\alpha_t}}{\beta_t} x_t+\frac{2 \sqrt{\bar{\alpha}_t}}{1-\bar{\alpha}_t} x_0\right) x_{t-1}+C\right]\right\}\right)
	\end{gathered}
\end{equation}

\begin{equation}
	\small
	\label{eq5}
	\begin{gathered}
		\tilde{\mu}=\frac{1}{\sqrt{\alpha_t}}\left(x_t-\frac{\beta_t}{\sqrt{1-\bar{\alpha}_t}} z_t\right) \\
		\widetilde{\beta_t}=\frac{1-\bar{\alpha}_{t-1}}{1-\bar{\alpha}_t} \beta_t
	\end{gathered}
\end{equation}

The negative log-likelihood function for the target distribution is computed as follows:

\begin{equation}
	\small
	E_{q\left(x_{0: T}\right)}\left[\log \frac{q\left(x_{1: T} \mid x_0\right)}{p_\theta\left(x_{0: T}\right)}\right] \geq-E_{q\left(x_0\right)} \log p_\theta\left(x_0\right)
\end{equation}

The current objective is to minimize the upper bound of the likelihood function. Combining equation \ref{eq5} with parameter renormalization yields a simplified form of loss function:

\begin{equation}
	\small
	\begin{aligned}
		& \min E_{q\left(x_{0: T}\right)}\left[\log \frac{q\left(x_{1: T} \mid x_0\right)}{p_\theta\left(x_{0: T}\right)}\right] \\
		\Rightarrow & \min D_{K L}\left(q\left(x_{t-1} \mid x_t, x_0\right) \| p_\theta\left(x_{t-1}, x_t\right)\right) \\
		\Rightarrow & \min E_q\left[\frac{1}{2 \sigma^2}\left\|\tilde{\mu}_t\left(x_t, x_0\right)-\mu_\theta\left(x_t, t\right)\right\|^2\right] \\
		\Rightarrow & \min E_{t, x_0, \epsilon}\left\|\epsilon-\epsilon_\theta\left(\sqrt{\bar{\alpha}_t} x_0+\sqrt{1-\bar{\alpha}_t} \epsilon, t\right)\right\|^2
	\end{aligned}
\end{equation}

Proceed to optimize the loss function and extract the training results for DDPM model $\epsilon_\theta$ in noise evaluation module.

\begin{figure}[!t]
	\centering
	\includegraphics[width=\linewidth]{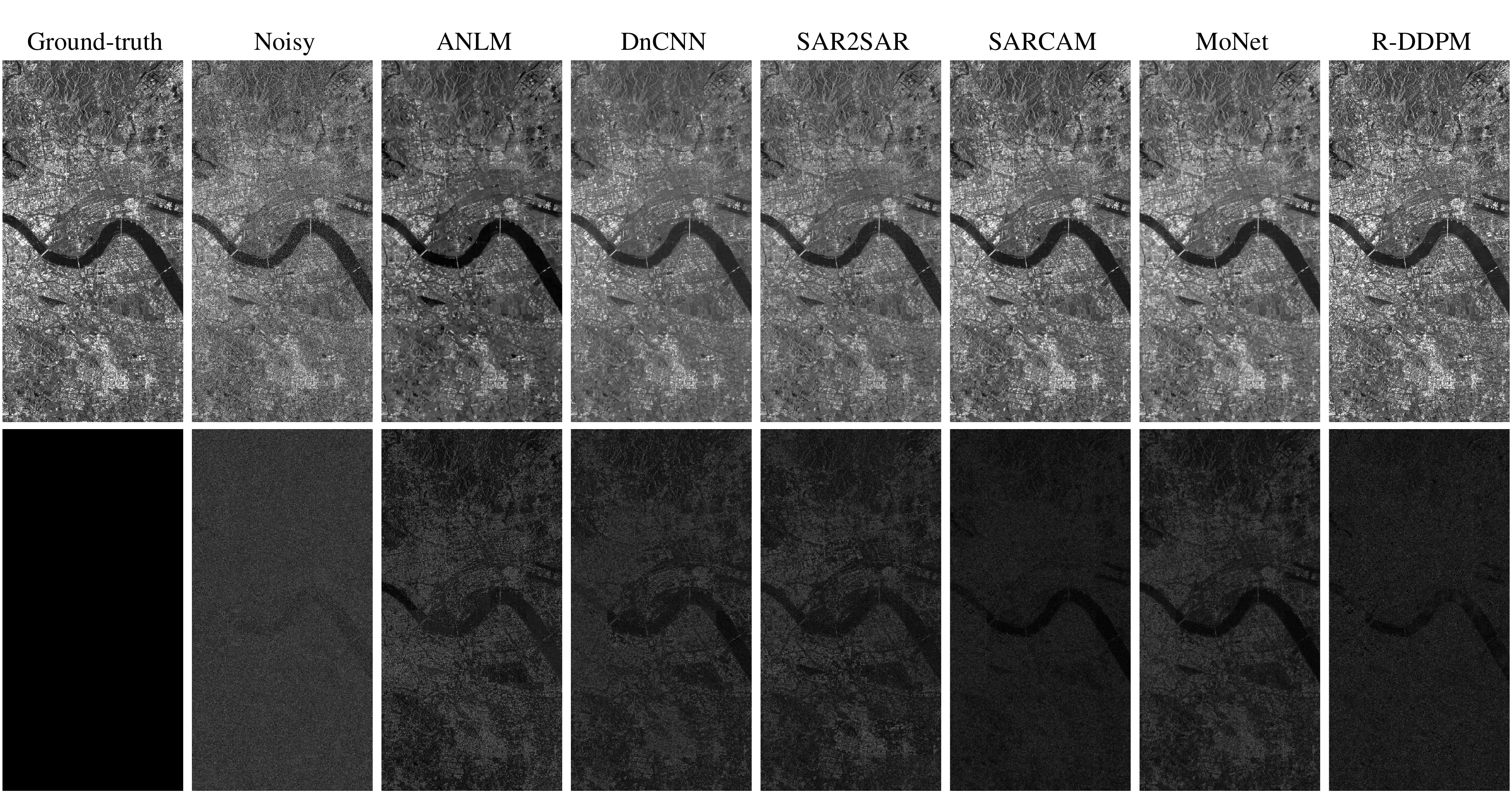}
	\caption{The demonstration of experiments on large-scale images. The first row showcases the despeckling reconstruction results, the second row displays the residual results.}
	\label{fig_3}
\end{figure}

\begin{figure}[!t]
	\centering
	\includegraphics[width=\linewidth]{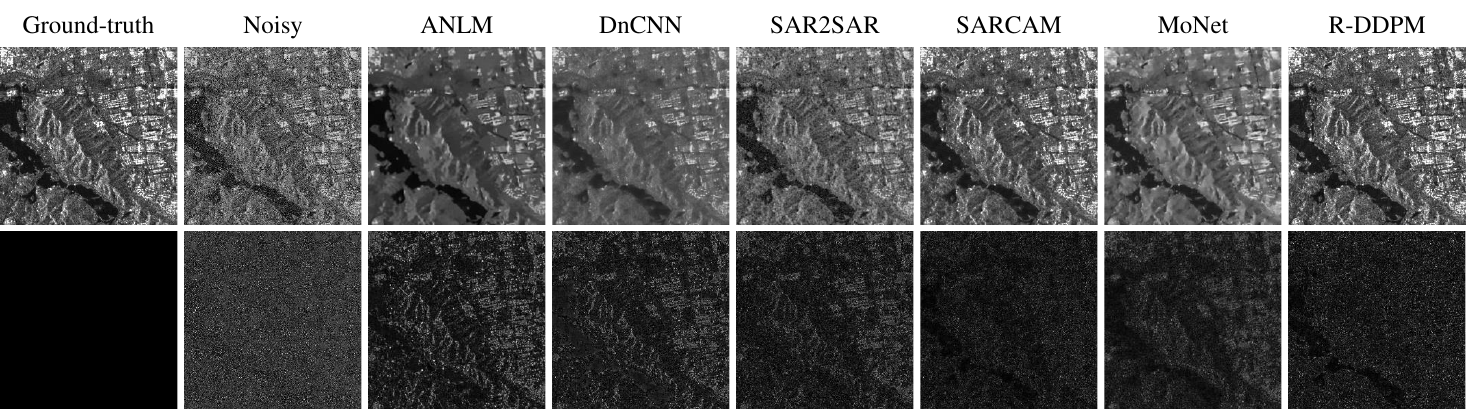}
	\caption{The demonstration of experiments on small-scale images. The first row showcases the despeckling reconstruction results, the second row displays the residual results.}
	\label{fig_4}
\end{figure}

\subsection{Conditional Diffusion}


In Conditional Diffusion \cite{dhariwal2021diffusion}, the condition $\tilde{x}$ is introduced during the reverse diffusion process, ensuring the generated $x_0$ adheres to the distribution of $\tilde{x}$. 

\begin{equation}
	p_\theta\left(x_{0: T} \mid \tilde{x}\right)=p\left(x_T\right) \prod_{t=1}^T p_\theta\left(x_{t-1} \mid x_t, \tilde{x}\right)
\end{equation}

Specifically, we achieve this conditionality by channel-wise concatenation of $x$ and $\tilde{x}$. employing a multi-channel format as the input for the reverse process.

\subsection{Regional Restoration}

Current deep learning-based methods often require inputs at fixed scales. Resizing input images can alter the noise levels, rendering pre-trained models ineffective. Utilizing block-based methods for despeckling image blocks and then assembling them will lead to edge artifacts. 

To handle SAR images of arbitrary scales, we adopt regional restoration to achieve artifact-free despeckling results. Figure 2 illustrates the implementation flow of regional restoration. We partition the noisy image $\tilde{x}$ and the inferred intermediate process $x_t$ into $K$ regions of size $m \times m$, denoting each image block as $\tilde{x}^k$ and $x_t^k$. Simultaneously, to mitigate edge artifacts resulting from region concatenation, we further divide each $m \times m$ region into smaller $R$ regions of size $n \times n$. For ease of computations, $n$ is set as an integer multiple of $m$, while the 'window' $x^k$ traverses the image regions with a step size of $m$.

For each region $k$, noise estimation function $\epsilon_\theta\left(x_t^k, \tilde{x}^k, t\right)$ is computed using the Diffusion model. Moreover, due to the window's movement causing partial overlap of regions $r$, these overlapped areas have their noise estimation functions calculated separately, and their average noise estimation is obtained through averaging. Embedding these individual regions into the original image space yields the overall noise estimation function $\epsilon_\theta^{\prime}\left(x_t, \tilde{x}_t, t, n, m\right)$. This culminates in the final sampling process:

\begin{equation}
	\small
	x_{t-1}=\frac{1}{\sqrt{\alpha_t}}\left(x_t-\frac{1-\alpha_t}{\sqrt{1-\bar{\alpha}_t}} \epsilon_\theta^{\prime}\left(x_t, \tilde{x}_t, t, n, m\right)\right)+\sigma_t I
\end{equation}

\section{Experimental Results}
\label{section:C}

\subsection{Experiment Settings}

\noindent \textbf{Dataset:} This study utilizes Sentinel-1 data \cite{geudtner2012sentinel} sampled near Shanghai, China. The polarization mode used is VV+VH. We add different levels of Gaussian noise to it to simulate the noise generated during the SAR image sampling process and form the datasets.

\noindent \textbf{Evaluation Metrics:} We conducted qualitative and quantitative evaluations on various despeckling methods. We sampled SAR images of different scales for denoising and obtained visual results. Additionally, we used common image restoration metrics, Peak Signal-to-Noise Ratio (PSNR), and Structural Similarity (SSIM) \cite{wang2004image} as quantitative evaluation metrics. In real-image experiments, as Ground-truth images are unavailable, Equivalent Number of Looks (ENL) and Edge Preservation Index (EPI) are employed as reference metrics.

\noindent \textbf{Implementation Details:} The Diffusion model employs a U-Net architecture with noise estimation as the loss function. The parameter $\beta$ $\in$ [0.0001, 0.02], with a time step set at 1000, a batch size of 4, and a learning rate of 0.00002. For region recovery, $m$ is set to 64, and $n$ is set to 16. All experiments are conducted using the PyTorch framework and executed on an RTX3080 GPU.

\begin{table}[!t]
	\caption{Quantitative (PSNR and SSIM) and subjective evaluation of synthetic images.}
	\label{tab1}
	\centering
	\scriptsize
	\renewcommand{\arraystretch}{1.3}
	\begin{tabular}{lccccc} 
		\toprule
		& \multicolumn{2}{c}{\textbf{Small-scale}} & \multicolumn{2}{c}{\textbf{Large-scale}} & \textbf{Sub.} \\
		\cmidrule(lr){2-3} \cmidrule(lr){4-5} 
		& \textbf{PSNR(dB)} & \textbf{SSIM(\%)} & \textbf{PSNR(dB)} & \textbf{SSIM(\%)} & \\
		\midrule
		ANLM\cite{xiao2020asymptotic} & 17.49 & 34.93 & 17.37 & 35.81 & 5.9 \\
		SAR2SAR\cite{dalsasso2021sar2sar} & 19.35 & 52.29 & 19.23 & 50.20 & 3.6 \\
		SARCAM\cite{ko2021sar} & 21.22 & \textbf{66.07} & 22.15 & 64.90 & \textbf{9.3} \\
		DnCNN\cite{zhang2017beyond} & 19.04 & 43.30 & 18.80 & 43.66 & 5.0 \\
		MoNet\cite{vitale2021analysis} & 19.64 & 49.09 & 19.49 & 48.22 & 5.4 \\
		R-DDPM & \textbf{21.93} & 64.38 & \textbf{22.39} & \textbf{65.02} & \textbf{9.3} \\
		\bottomrule
	\end{tabular}
\end{table}

\subsection{Comparison with State-of-the-Art}


In this section, our method is compared with ANLM \cite{xiao2020asymptotic}, MoNet \cite{vitale2021analysis}, SAR2SAR \cite{dalsasso2021sar2sar}, SARCAM \cite{ko2021sar}, and DnCNN \cite{zhang2017beyond}. We retrained SAR2SAR, SARCAM, and DnCNN for despeckling within their frameworks. MoNet utilized pre-trained weights for image despeckling. ANLM is a transform-based method where we adjusted the optimal parameters to achieve despeckling results.


\noindent \textbf{Qualitative Experiments:} R-DDPM directly performs despeckling on the 2560 $\times$ 5120 image, while other methods despeckle 50 individual 512 $\times$ 512 patches before assembling the results. Figure \ref{fig_3} presents the experimental results of despeckling large-scale SAR images. R-DDPM and SARCAM achieved better results, whereas DnCNN, SAR2SAR, and MoNet inevitably resulted in blurring, causing various degrees of loss in image details. ANLM, as a traditional despeckling method, overly smoothed certain areas, leading to the loss of some details. Relying on the robust generative capacity of the Diffusion model and region recovery sampling, R-DDPM effectively avoids edge artifacts and image blurring.

Small-scale despeckling experiment: Randomly selected portions of 512 $\times$ 512 samples are tested for despeckling, as shown in Figure \ref{fig_4}. The experiments indicate that R-DDPM retains more details compared to other methods, significantly enhancing image contrast and clarity, thereby validating the effectiveness of the proposed method.

For qualitative assessment, we introduced subjective evaluations to determine image quality. Using Ground-truth as the reference, we collected subjective ratings from 22 individuals for various despeckling methods (rated on a scale of 1-10, where higher scores indicate better despeckling effects). The last column of Table \ref{tab1} displays the results of the subjective evaluation.

\begin{figure}[!t]
	\centering
	\includegraphics[width=\linewidth]{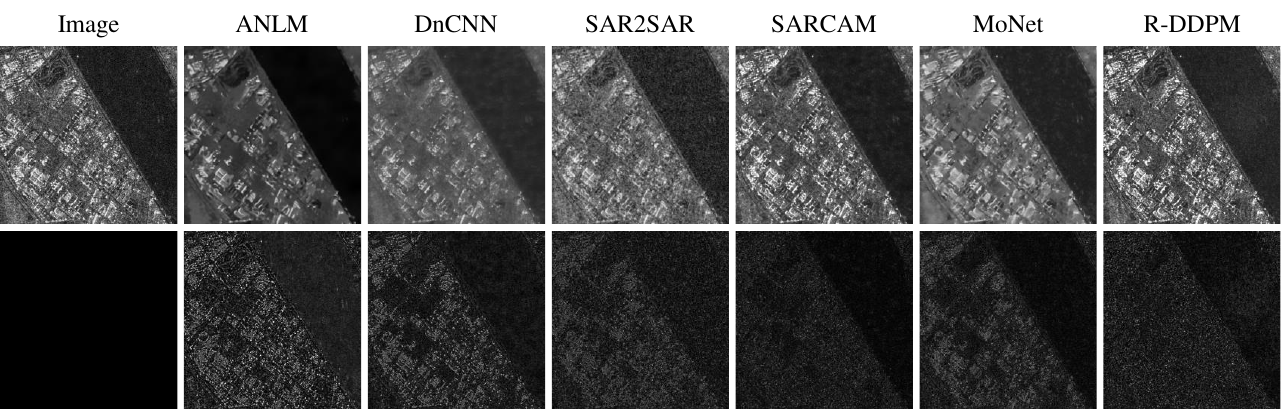}
	\caption{The demonstration of experiments on real SAR images, where the first row showcases the despeckling reconstruction results, the second row displays the residual results.}
	\label{fig_5}
\end{figure}

\begin{table}[!t]
	\caption{Quantitative evaluation (ENL, EPI) of real images.}
	\label{tab2}
	\centering
	\scriptsize
	\renewcommand{\arraystretch}{1.3}
	\begin{tabular}{p{2cm} >{\centering\arraybackslash}m{2cm} >{\centering\arraybackslash}m{2cm}}
		\toprule
		& \textbf{ENL} & \textbf{EPI} \\
		\midrule
		ANLM\cite{xiao2020asymptotic} & 51.21 & 6.418 \\
		SAR2SAR\cite{dalsasso2021sar2sar} & 98.69 & 8.562 \\
		SARCAM\cite{ko2021sar} & 131.47 & 5.220 \\
		DnCNN\cite{zhang2017beyond} & 59.68 & \textbf{13.007} \\
		MoNet\cite{vitale2021analysis} & 60.72 & 10.067  \\
		R-DDPM & \textbf{179.70} & 6.126 \\
		\bottomrule
	\end{tabular}
\end{table}

\noindent \textbf{Quantitative Experiments:} In this section, we assessed the performance of different despeckling methods by evaluating their PSNR and SSIM against the Ground-truth, a measure aimed at gauging their effectiveness. Our initial tests were conducted on small-scale image data, utilizing 512 $\times$ 512 image blocks as inputs to ensure a fair and consistent comparison across methods. Simultaneously, we included a sample of large-scale image data, which presented a more formidable challenge for each method's despeckling capabilities. Upon reevaluation of all methods using this dataset, the findings summarized in Table 1 showcase the experimental outcomes under both small and large-scale conditions.

The results of these experiments highlight R-DDPM's ability to effectively eliminate noise while retaining intricate texture details within SAR images, exhibiting superior performance in both PSNR and SSIM metrics. Notably, during the despeckling process of larger-scale SAR images, the utilization of the region recovery sampling ensures a greater level of image consistency. SARCAM also exhibited commendable performance in our tests. Conversely, alternative methods yielded images with increased blurriness and loss of finer details, showcasing varying degrees of degradation in metrics when handling larger-scale images.

\noindent \textbf{Real Image Experiments:} In this section, we employed the model pre-trained on synthetic images for real SAR image despeckling. Given the absence of Ground-truth images in real-image experiments, SSIM and PSNR were not used as evaluation metrics. Instead, we opted for ENL and EPI as substitutes to gauge the noise intensity in denoised images. It's important to note that these metrics serve as references; while they partly measure the noise level in images, they do not encompass parameters like image clarity. Therefore, they are used solely as reference metrics. Table \ref{tab2} presents the ENL and EPI results.

Figure \ref{fig_5} illustrates the qualitative comparison of results from real-image experiments. Due to the introduction of varying degrees of Gaussian noise during the training process in synthetic image experiments, R-DDPM exhibits a certain adaptability to noise, retaining more image details in real-image experiments compared to other methods, which still display tendencies toward excessive smoothing. Quantitatively, according to the results in Table \ref{tab2}, R-DDPM achieves the best performance in terms of the ENL metric, while DnCNN records the highest EPI metric.

%

\section{Conclusion}

This paper introduces a Regional Denoising Diffusion Probabilistic Model, R-DDPM, which achieves despeckling of SAR images with arbitrary scale inputs and low resource consumption. The model segments images into specified-sized regions, constructs a dataset for the diffusion training process, and utilizes overlapping region sampling in the reverse process to guide inverse diffusion, thereby achieving high-quality SAR image despeckling without edge artifacts. This lightweight model requires less than 12GB of GPU memory for training and is easily deployable. Experimental results demonstrate that R-DDPM's despeckling capability surpasses the current state-of-the-art methods.

\bibliographystyle{IEEEbib}
\bibliography{strings,refs}

\end{document}